\documentclass[11pt,letterpaper]{article}
\usepackage[hyperref]{emnlp2018}
\usepackage{times}
\usepackage{latexsym}
\pdfoutput=1

\usepackage{graphicx}
\graphicspath{ {./rsc/} }
\usepackage{tabularx}

\usepackage{url}
\usepackage{amssymb}

\title{Comparative Studies of Detecting Abusive Language on Twitter}

\aclfinalcopy

\author{Younghun Lee\thanks{* Equal contribution.}~~~~~~~Seunghyun Yoon\footnotemark[1]~~~~~~~Kyomin Jung \\
  Dept. of Electrical and Computer Engineering \\
  Seoul National University, Seoul, Korea \\
  {\tt younggnse@gmail.com~\{mysmilesh,kjung\}@snu.ac.kr} }

\date{}

\begin{document}
\maketitle
\begin{abstract} 
The context-dependent nature of online aggression makes annotating large collections of data extremely difficult.
Previously studied datasets in abusive language detection have been insufficient in size to efficiently train deep learning models.
Recently, \emph{Hate and Abusive Speech on Twitter}, a dataset much greater in size and reliability, has been released.
However, this dataset has not been comprehensively studied to its potential.
In this paper, we conduct the first comparative study of various learning models on \emph{Hate and Abusive Speech on Twitter}, and discuss the possibility of using additional features and context data for improvements.
Experimental results show that bidirectional GRU networks trained on word-level features, with Latent Topic Clustering modules, is the most accurate model scoring 0.805 F1.
\end{abstract}

\section{Introduction}
Abusive language refers to any type of insult, vulgarity, or profanity that debases the target; it also can be anything that causes aggravation~\cite{spertus1997smokey,schmidt2017survey}.
Abusive language is often reframed as, but not limited to, offensive language~\cite{razavi2010offensive}, cyberbullying~\cite{xu2012learning}, othering language~\cite{burnap2014hate}, and hate speech~\cite{djuric2015hate}.

Recently, an increasing number of users have been subjected to harassment, or have witnessed offensive behaviors online~\cite{Duggan:17}. Major social media companies (i.e. Facebook, Twitter) have utilized multiple resources---artificial intelligence, human reviewers, user reporting processes, etc.---in effort to censor offensive language, yet it seems nearly impossible to successfully resolve the issue~\cite{Robertson:17,Musaddique:17}.

The major reason of the failure in abusive language detection comes from its subjectivity and context-dependent characteristics~\cite{chatzakou2017mean}.
For instance, a message can be regarded as harmless on its own, but when taking previous threads into account it may be seen as abusive, and vice versa.
This aspect makes detecting abusive language extremely laborious even for human annotators; therefore it is difficult to build a large and reliable dataset~\cite{founta2018large}.

Previously, datasets openly available in abusive language detection research on Twitter ranged from 10K to 35K in size~\cite{chatzakou2017mean,golbeck2017large}.
This quantity is not sufficient to train the significant number of parameters in deep learning models.
Due to this reason, these datasets have been mainly studied by traditional machine learning methods.
Most recently, Founta et al.~\shortcite{founta2018large} introduced \emph{Hate and Abusive Speech on Twitter}, a dataset containing 100K tweets with cross-validated labels.
Although this corpus has great potential in training deep models with its significant size, there are no baseline reports to date.

This paper investigates the efficacy of different learning models in detecting abusive language. We compare accuracy using the most frequently studied machine learning classifiers as well as recent neural network models.\footnote{The code can be found at: \url{https://github.com/younggns/comparative-abusive-lang}} Reliable baseline results are presented with the first comparative study on this dataset.
Additionally, we demonstrate the effect of different features and variants,
and describe the possibility for further improvements with the use of ensemble models.

\section{Related Work}
The research community introduced various approaches on abusive language detection. 
Razavi et al.~\shortcite{razavi2010offensive} applied Na\"ive Bayes, and Warner and Hirschberg~\shortcite{warner2012detecting} used Support Vector Machine (SVM), both with word-level features to classify offensive language. 
Xiang et al.~\shortcite{xiang2012detecting} generated topic distributions with Latent Dirichlet Allocation~\cite{blei2003latent}, also using word-level features in order to classify offensive tweets.

More recently, distributed word representations and neural network models have been widely applied for abusive language detection.
Djuric et al.~\shortcite{djuric2015hate} used the Continuous Bag Of Words model with paragraph2vec algorithm~\cite{le2014distributed} to more accurately detect hate speech than that of the plain Bag Of Words models.
Badjatiya et al.~\shortcite{badjatiya2017deep} implemented Gradient Boosted Decision Trees classifiers using word representations trained by deep learning models.
Other researchers have investigated character-level representations and their effectiveness compared to word-level representations~\cite{mehdad2016characters,park2017one}.

As traditional machine learning methods have relied on feature engineering, (i.e. n-grams, POS tags, user information)~\cite{schmidt2017survey}, researchers have proposed neural-based models with the advent of larger datasets.
Convolutional Neural Networks and Recurrent Neural Networks have been applied to detect abusive language, and they have outperformed traditional machine learning classifiers such as Logistic Regression and SVM~\cite{park2017one,badjatiya2017deep}.
However, there are no studies investigating the efficiency of neural models with large-scale datasets over 100K.

\section{Methodology}
This section illustrates our implementations on traditional machine learning classifiers and neural network based models in detail. 
Furthermore, we describe additional features and variant models investigated.

\subsection{Traditional Machine Learning Models}
We implement five feature engineering based machine learning classifiers that are most often used for abusive language detection. 
In data preprocessing, text sequences are converted into Bag Of Words (BOW) representations, and normalized with Term Frequency-Inverse Document Frequency (TF-IDF) values.
We experiment with word-level features using n-grams ranging from 1 to 3, and character-level features from 3 to 8-grams.
Each classifier is implemented with the following specifications:
\\

\noindent
\textbf{Na\"ive Bayes (NB)}: Multinomial NB with additive smoothing constant 1\\
\noindent
\textbf{Logistic Regression (LR)}: Linear LR with L2 regularization constant 1 and limited-memory BFGS optimization\\
\noindent
\textbf{Support Vector Machine (SVM)}: Linear SVM with L2 regularization constant 1 and logistic loss function\\
\noindent
\textbf{Random Forests (RF)}: Averaging probabilistic predictions of 10 randomized decision trees\\
\noindent
\textbf{Gradient Boosted Trees (GBT)}: Tree boosting with learning rate 1 and logistic loss function

\subsection{Neural Network based Models}
Along with traditional machine learning approaches, we investigate neural network based models to evaluate their efficacy within a larger dataset.
In particular, we explore Convolutional Neural Networks (CNN), Recurrent Neural Networks (RNN), and their variant models. 
A pre-trained GloVe~\cite{pennington2014glove} representation is used for word-level features.\\

\noindent
\textbf{CNN}: 
We adopt Kim's~\shortcite{kim2014convolutional} implementation as the baseline.
The word-level CNN models have 3 convolutional filters of different sizes [1,2,3] with ReLU activation, and a max-pooling layer.
For the character-level CNN, we use 6 convolutional filters of various sizes [3,4,5,6,7,8], then add max-pooling layers followed by 1 fully-connected layer with a dimension of 1024.

Park and Fung~\shortcite{park2017one} proposed a HybridCNN model which outperformed both word-level and character-level CNNs in abusive language detection.
In order to evaluate the HybridCNN for this dataset, we concatenate the output of max-pooled layers from word-level and character-level CNN, and feed this vector to a fully-connected layer in order to predict the output.

All three CNN models (word-level, character-level, and hybrid) use cross entropy with softmax as their loss function and Adam~\cite{kingma2014adam} as the optimizer.\\

\noindent
\textbf{RNN}: 
We use bidirectional RNN~\cite{schuster1997bidirectional} as the baseline, implementing a GRU~\cite{cho2014learning} cell for each recurrent unit.
From extensive parameter-search experiments, we chose 1 encoding layer with 50 dimensional hidden states and an input dropout probability of 0.3.
The RNN models use cross entropy with sigmoid as their loss function and Adam as the optimizer.

For a possible improvement, we apply a self-matching attention mechanism on RNN baseline models~\cite{wang2017gated} so that they may better understand the data by retrieving text sequences twice.
We also investigate a recently introduced method, Latent Topic Clustering (LTC)~\cite{yoon2018learning}.
The LTC method extracts latent topic information from the hidden states of RNN, and uses it for additional information in classifying the text data.

\subsection{Feature Extension}
While manually analyzing the raw dataset, we noticed that looking at the tweet one has replied to or has quoted, provides significant contextual information.
We call these, \emph{``context tweets"}.
As humans can better understand a tweet with the reference of its context, our assumption is that computers also benefit from taking context tweets into account in detecting abusive language.

As shown in the examples below, (2) is labeled abusive due to the use of vulgar language. 
However, the intention of the user can be better understood with its context tweet (1).\\

\begin{small}
\noindent
(1) I hate when I'm sitting in front of the bus and somebody with a wheelchair get on.\\
\noindent
\rotatebox[origin=c]{180}{$\Lsh$} (2) I hate it when I'm trying to board a bus and there's already an as**ole on it.\\
\end{small}

Similarly, context tweet (3) is important in understanding the abusive tweet (4), especially in identifying the target of the malice.\\

\begin{small}
\noindent
(3) Survivors of \#Syria Gas Attack Recount `a Cruel Scene'.\\
\noindent
\rotatebox[origin=c]{180}{$\Lsh$} (4) Who the HELL is ``LIKE" ING this post? Sick people....\\
\end{small}

Huang et al.~\shortcite{huang2016modeling} used several attributes of context tweets for sentiment analysis in order to improve the baseline LSTM model.
However, their approach was limited because the meta-information they focused on---author information, conversation type, use of the same hashtags or emojis---are all highly dependent on data.

In order to avoid data dependency, text sequences of context tweets are directly used as an additional feature of neural network models.
We use the same baseline model to convert context tweets to vectors, then concatenate these vectors with outputs of their corresponding labeled tweets.
More specifically, we concatenate max-pooled layers of context and labeled tweets for the CNN baseline model. As for RNN, the last hidden states of context and labeled tweets are concatenated.

\section{Experiments} 

\subsection{Dataset} 
\begin{table}[t!]
\centering
\scalebox{0.85}{
\begin{tabular}{|r||c|c|c|c|}
\hline
Labels & Normal & Spam & Hateful & Abusive\\
\hline
Number & 42,932 & 9,757 & 3,100 & 15,115 \\
(\%) & (60.5) & (13.8) & (4.4) & (21.3) \\
\hline
\end{tabular}
}
\caption{Label distribution of crawled tweets}
\label{table_crawledTweets_labels} 
\end{table}
~\emph{Hate and Abusive Speech on Twitter}~\cite{founta2018large} classifies tweets into 4 labels, \emph{``normal"}, \emph{``spam"}, \emph{``hateful"} and \emph{``abusive"}. We were only able to crawl 70,904 tweets out of 99,996 tweet IDs, mainly because the tweet was deleted or the user account had been suspended. Table~\ref{table_crawledTweets_labels} shows the distribution of labels of the crawled data.

\subsection{Data Preprocessing}
In the data preprocessing steps, user IDs, URLs, and frequently used emojis are replaced as special tokens. Since hashtags tend to have a high correlation with the content of the tweet~\cite{lehmann2012dynamical}, we use a segmentation library\footnote{WordSegment module description page: \url{https://pypi.org/project/wordsegment/}}~\cite{segaran:09} for hashtags to extract more information.

For character-level representations, we apply the method Zhang et al.~\shortcite{zhang2015character} proposed. Tweets are transformed into one-hot encoded vectors using 70 character dimensions---26 lower-cased alphabets, 10 digits, and 34 special characters including whitespace.

\begin{table*}[t!]
\centering
\small
\scalebox{0.87}{
\renewcommand{\arraystretch}{1.2}
\begin{tabular}{|r|ccc|ccc|ccc|ccc||ccc|}

\hline  
& \multicolumn{3}{c|}{\bf Normal} & \multicolumn{3}{c|}{\bf Spam} & \multicolumn{3}{c|}{\bf Hateful} & \multicolumn{3}{c||}{\bf Abusive} & \multicolumn{3}{c|}{\bf Total} \\
Model & Prec. & Rec. & F1 & Prec. & Rec. & F1 & Prec. & Rec. & F1 & Prec. & Rec. & F1 & Prec. & Rec. & F1 \\
\hline
NB (word) & .776 & .916 & .840 & .573 & .378 & .456 & .502 & .034 & .063 & .828 & .744 & .784 & .747 & .767 & .741 \\
NB (char) & .827 & .805 & .815 & .467 & .\textbf{609} & .528 & .452 & .061 & .107 & .788 & .832 & .803 & .752 & .751 & .744 \\
\hline
LR (word) & .807 & .933 & .865 & .616 & .365 & .458 & .620 & .161 & .254 & .868 & .844 & .856 & .786 & .802 & .780 \\
LR (char) & .808 & .934 & .866 & .618 & .363 & .457 & .636 & .183 & .283 & .\textbf{873} & .848 & .860 & .788 & .804 & .783 \\
\hline
SVM (word) & .757 & .967 & .850 & .\textbf{678} & .190 & .296 & .\textbf{836} & .034 & .065 & .865 & .757 & .807 & .773 & .775 & .730 \\
SVM (char) & .763 & .\textbf{968} & .853 & .\textbf{680} & .198 & .306 & .\textbf{805} & .070 & .129 & .\textbf{876} & .775 & .822 & .778 & .781 & .740 \\
\hline
RF (word) & .776 & .945 & .853 & .581 & .213 & .311 & .556 & .109 & .182 & .852 & .819 & .835 & .757 & .781 & .745 \\
RF (char) & .793 & .934 & .857 & .568 & .252 & .349 & .563 & .150 & .236 & .853 & .856 & .854 & .765 & .789 & .760 \\
\hline
GBT (word) & .806 & .921 & .860 & .581 & .320 & .413 & .506 & .194 & .279 & .854 & .863 & .858 & .772 & .794 & .773 \\
GBT (char) & .807 & .913 & .857 & .560 & .346 & .428 & .472 & .187 & .267 & .859 & .859 & .859 & .770 & .791 & .772 \\
\hline
\hline

CNN (word) & .822 & .925 & .870 & .625 & .323 & .418 & .563 & .182 & .263 & .846 & .916 & .879 & .789 & .808 & .783 \\
CNN (char) & .784 & .946 & .857 & .604 & .180 & .264 & .663 & .124 & .204 & .848 & .864 & .856 & .768 & .787 & .747 \\
CNN (hybrid) & .820 & .926 & .869 & .616 & .322 & .407 & .628 & .180 & .265 & .853 & .910 & .880 & .790 & .807 & .781 \\
\hline

RNN~(word) & .856 & .887 & .870 & .589 & .514 & .547 & .577 & .194 & .287 & .844 & \textbf{.934} & .\textbf{887} & .\textbf{804} & .\textbf{815} & .\textbf{804} \\
RNN~(char) & .606 & .\textbf{999} & .754 & .000 & .000 & .000 & .000 & .000 & .000 & .000 & .000 & .000 & .367 & .605 & .457 \\
RNN-attn (word) & .846 & .898 & .\textbf{872} & .593 & .469 & .520 & .579 & .194 & .283 & .849 & .925 & .886 & .800 & .814 & .800 \\

RNN-LTC (word) & .\textbf{857} & .884 & .\textbf{871} & .583 & .525 & .\textbf{551} & .564 & .\textbf{210} & .\textbf{302} & .846 & .932 & .\textbf{887} & .\textbf{804} & .\textbf{815} & .\textbf{805} \\
\hline
\hline
CNN (w/context) & .828 & .910 & .867 & .609 & .341 & .429 & .505 & .\textbf{246} & .\textbf{309} & .840 & .914 & .875 & .786 & .804 & .784 \\
\hline
RNN (w/context) & .\textbf{858} & .880 & .869 & .577 & \textbf{.527} & .\textbf{549} & .534 & .175 & .256 & .840 & .\textbf{937} & .885 & .801 & .813 & .801 \\
\hline

\end{tabular}
}

\caption{
Experimental results of learning models and their variants, followed by the context tweet models. The top 2 scores are marked as bold for each metric.
}
\label{table_performance}
\end{table*}
\subsection{Training and Evaluation}
In training the feature engineering based machine learning classifiers, we truncate vector representations according to the TF-IDF values (the top 14,000 and 53,000 for word-level and character-level representations, respectively) to avoid overfitting. For neural network models, words that appear only once are replaced as unknown tokens. 

Since the dataset used is not split into train, development, and test sets, we perform 10-fold cross validation, obtaining the average of 5 tries; we divide the dataset randomly by a ratio of 85:5:10, respectively.
In order to evaluate the overall performance, we calculate the weighted average of precision, recall, and F1 scores of all four labels, ``\emph{normal}'', ``\emph{spam}'', ``\emph{hateful}'', and ``\emph{abusive}''.

\subsection{Empirical Results}

As shown in Table~\ref{table_performance}, neural network models are more accurate than feature engineering based models (i.e. NB, SVM, etc.) except for the LR model---the best LR model has the same F1 score as the best CNN model. 

Among traditional machine learning models, the most accurate in classifying abusive language is the LR model followed by ensemble models such as GBT and RF. Character-level representations improve F1 scores of SVM and RF classifiers, but they have no positive effect on other models.

For neural network models, RNN with LTC modules have the highest accuracy score, but there are no significant improvements from its baseline model and its attention-added model. Similarly, HybridCNN does not improve the baseline CNN model. For both CNN and RNN models, character-level features significantly decrease the accuracy of classification.

The use of context tweets generally have little effect on baseline models, however they noticeably improve the scores of several metrics. For instance, CNN with context tweets score the highest recall and F1 for ``\textit{hateful}" labels, and RNN models with context tweets have the highest recall for ``\textit{abusive}" tweets.

\section{Discussion and Conclusion}
While character-level features are known to improve the accuracy of neural network models~\cite{badjatiya2017deep}, they reduce classification accuracy for \emph{Hate and Abusive Speech on Twitter}. We conclude this is because of the lack of labeled data as well as the significant imbalance among the different labels. Unlike neural network models, character-level features in traditional machine learning classifiers have positive results because we have trained the models only with the most significant character elements using TF-IDF values.

Variants of neural network models also suffer from data insufficiency. 
However, these models show positive performances on \emph{``spam"} (14\%) and \emph{``hateful"} (4\%) tweets---the lower distributed labels.
The highest F1 score for \emph{``spam"} is from the RNN-LTC model (0.551), and the highest for \emph{``hateful"} is CNN with context tweets (0.309).
Since each variant model excels in different metrics, we expect to see additional improvements with the use of ensemble models of these variants in future works.

In this paper, we report the baseline accuracy of different learning models as well as their variants on the recently introduced dataset, \emph{Hate and Abusive Speech on Twitter}. 
Experimental results show that bidirectional GRU networks with LTC provide the most accurate results in detecting abusive language. 
Additionally, we present the possibility of using ensemble models of variant models and features for further improvements.

\section*{Acknowledgments}
K. Jung is with the Department of Electrical and Computer Engineering, ASRI, Seoul National University, Seoul, Korea. 
This work was supported by the National Research Foundation of Korea (NRF) funded by the Korea government (MSIT) (No. 2016M3C4A7952632), the Technology Innovation Program (10073144) funded by the Ministry of Trade, Industry \& Energy (MOTIE, Korea). 

We would also like to thank Yongkeun Hwang and Ji Ho Park for helpful discussions and their valuable insights.

\bibliography{emnlp2018}
\bibliographystyle{acl_natbib_nourl}

\end{document}